%% file: ae.tex
\pdfoutput=1

\documentclass[11pt,a4paper]{article}
\usepackage[]{emnlp2022}

\usepackage{makecell}
\usepackage{times}
\usepackage{latexsym}
\usepackage{tabularx}
\usepackage{array}
\usepackage{multirow}
\usepackage{siunitx}
\usepackage{booktabs}
\usepackage[pdftex]{graphicx}
\usepackage{enumitem}
\usepackage{xspace}
\usepackage{caption}
\usepackage{subcaption}
\usepackage[english]{babel}

\usepackage{microtype}

\usepackage{amsthm}

\theoremstyle{definition}
\newtheorem{definition}{Definition}[section]

\newcommand{\abr}[1]{\textsc{#1}}
\newcommand{\squad}{SQuAD\xspace}

\newcommand{\fone}{\texorpdfstring{F\textsubscript{1}\xspace}{F1\xspace}}
\newcommand{\ourscore}{BEM\xspace}

\title{Tomayto, Tomahto. Beyond Token-level Answer Equivalence \\ for Question Answering Evaluation}

\author{
  Jannis Bulian\quad Christian Buck\quad Wojciech Gajewski\\{\bf Benjamin B\"{o}rschinger}\quad {\bf Tal Schuster}\\
  \\\vspace{0.5cm}
  Google Research\\
  \texttt{\{jbulian, cbuck, wgaj, bboerschinger, talschuster\}@google.com}
  }

\date{}

\begin{document}
\maketitle

\input{sections/00-abstract}
\input{sections/10-intro}

\input{sections/20-f1-em-issues}
\input{sections/30-data}
\input{sections/40-model}

\input{sections/41-conformal}

\input{sections/69-related}
\input{sections/70-conclusion}

\section*{Acknowledgments}
We thank Adam Fisch for helpful feedback on the conformal prediction experiments,
and Costanza Conforti for helping us to add the dataset to TFDS.

\input{sections/80-ethics}
\input{sections/81-limitations}

\bibliography{anthology,pragma}
\bibliographystyle{acl_natbib}

\clearpage
\section*{Appendix}
\appendix
\input{sections/90-appendix}
\input{sections/98-data-stats}
\input{sections/99-a-bert-details}

\input{sections/99-b-conf-details}

\end{document}

%% file: sections/00-abstract.tex
\begin{abstract}

The predictions of question answering (QA) systems are typically evaluated against manually annotated finite sets of one or more answers. This
leads to a coverage limitation that results in underestimating the true performance of systems, and is typically addressed by extending over exact match (EM) with predefined rules or with the token-level \fone measure. 
In this paper, we present the first systematic conceptual and data-driven analysis to examine the shortcomings of token-level equivalence measures.\\
To this end, we define the asymmetric notion of answer equivalence (AE), accepting answers that are \emph{equivalent to or improve over the reference}, and publish over 23k human judgments for candidates produced by multiple QA systems on \squad.\footnote{Dataset and model information available at:
\url{https://github.com/google-research-datasets/answer-equivalence-dataset}.\\The BEM model can be found at: \url{https://tfhub.dev/google/answer_equivalence/bem/1}.}
~Through a careful analysis of this data, we reveal and quantify several concrete limitations of the \fone measure, such as a false impression of graduality, or missing dependence on the question.\\
Since collecting AE annotations for each evaluated model is expensive, we learn a BERT matching (\ourscore) measure to approximate this task. Being a simpler task than QA, we find BEM to provide significantly better AE approximations than \fone, and to more accurately reflect the performance of systems.\\
Finally, we demonstrate the practical utility of AE and \ourscore on the concrete application of minimal accurate prediction sets, reducing the number of required answers by up to $\times2.6$.

\end{abstract}

%% file: sections/10-intro.tex
\section{Introduction}

Automatically assessing the answers given by question answering (QA) systems for correctness can be nontrivial.
This was already recognized in the early large-scale QA evaluation work of \citet{vorhees2000}, leading them to suggest string-matching patterns for approximation, yet recognizing their limitations:

\begin{quote}
\emph{
 ``it is quite difﬁcult to determine automatically whether the difference between a new string and a judged string is significant with respect to the correctness of the answer.''}
\end{quote}

Despite the early recognition of the importance and difficulty of evaluation for question answering, surprisingly little progress has been made.  As of today, QA research in the NLP community relies almost exclusively on two token-level metrics, Exact Match (EM) and Token \fone\ (\fone).
Unfortunately, 
both fall short of capturing the difference between \emph{significant and insignificant span differences}.
In Section~\ref{sec:issues}, we first document these %
limitations, providing a detailed analysis and examples from the \squad dataset.
Moreover, we extend this analysis with our new findings: the inherent 
asymmetry of the task and the reliance on the question,
also providing examples from the \squad dataset. We also  identify cases where the context is important.

One obvious limitation of token-level measures is their direct dependence on the diversity of the reference answers collected for the dataset~\citep{chen-etal-2019-evaluating}. While this could be addressed by extending the annotations, this is both expensive and has diminishing returns as the true collection of all correct answers might be large. In contrast, we focus on improving the equivalence measure beyond token matching
and thereby increase the answer inclusiveness over any reference set.

To facilitate research on this issue, we introduce
a well-defined \emph{Answer Equivalence} (AE) task along with the release of a new dataset (Section~\ref{sec:data}).
We collect human annotations on
\squad\ examples, comparing gold answers with model predictions.

We utilize this data in several ways. First, we use the human judgments
to better understand how well \fone\ and EM capture
equivalence in the machine reading setting. We demonstrate that
(1) both metrics underestimate the quality of candidate answers,
(2) are highly reliant on the number of available references, and
(3) that \fone\ gives the false impression of a gradual
rating, while really there is a similar
ratio of equivalent to non-equivalent answers for non-zero \fone.

We further propose to learn a measure that approximates the AE relations (Section~\ref{sec:predicting_AE}). To study this, we introduce BERT matching (\ourscore), that uses a BERT model trained on our data to predict equivalence. Through multiple experiments with three QA systems, we show that \ourscore better correlates with human judgments, compared to $F_1$.

Finally, we demonstrate the utility of AE and BEM on the concrete application of returning small and accurate answers sets (Section~\ref{sec:conformal_main}). 
Building on the \emph{expanded admission conformal prediction} framework~\citep{fisch2021efficient}, we show that AE and \ourscore significantly reduce the number of answers in the prediction sets while including a correct answer with arbitrarily high probability. Thanks to its simplicity, BEM can easily replace $F_1$ in many other applications that leverage QA predictions~\cite[e.g.,][]{honovich-etal-2021-q2, honovich-etal-2022-true-evaluating, eyal-etal-2019-question, alex2021qafacteval,schuster-etal-2021-get}.

Our main contributions include:%
\begin{itemize}[leftmargin=*]
    \itemsep0em 
    \item Defining the answer equivalence (AE) task.
    \item Releasing a large dataset with AE annotations.
    \item A data-driven analysis of the shortcomings of EM and $F_1$ as evaluation measures for QA.
    \item A learned AE measure (BEM) that better correlates with human judgments and enables practical improvements for QA-based applications.
\end{itemize}

%% file: sections/20-f1-em-issues.tex
\section{Common token-level metrics: EM \& \fone}
\label{sec:issues}

\paragraph{Notation.} We refer to the ``gold'' answers from humans as \emph{reference answers}. Each question is paired with a set of one or more gold answers.
We refer to the predicted answers produced by question answering models as \emph{candidate answers}. %

The most popular metrics are Token $F_1$ ($F_1$) and Exact Match (EM), defined as follows:

\begin{definition}[Exact Match]
Given a set of references $A_r$ and a candidate answer
$c$, $c$ is an \emph{exact match} for $A_r$ iff $c \in A_r$.
\end{definition}

\begin{definition}[Token \fone]
Given a set of references $A_r$, a candidate answer
$c$ and a tokenization function $t$, the \emph{Token \fone\
of $c$ with respect to $A_r$ and $t$} is the maximum over the token-wise \fone\ scores between $c$ and each $a \in A_r$, i.e.:
\[
\max_{a \in A_r}~ 2\left({\frac{|t(a)|}{|t(a) \cap t(c)|} + \frac{|t(c)|}{|t(a) \cap t(c)|}}\right)^{-1}
\]
\end{definition}

It is also common to remove stop words and punctuation before computing either EM or \fone.

\subsection{Limitations of  EM and \fone}

We briefly recount and provide examples for known short-comings.
EM and \fone\ imperfectly capture the answer equality and can over-, or underestimate the performance of models \citep{kocmi2021ship,gembenchmark2021,chen-etal-2019-evaluating,chen-et-al-2020}.
\paragraph{Strictness.} EM is often too strict, especially when only a few gold answers are available. Consider Example~1 in Table \ref{tab:examples}: even though the difference is a minor surface variation, the candidate receives an EM-score of 0.0 -- and also an \fone\ of $0.0$ (unless further tokenization is done).
\paragraph{Granularity.}
Both EM and \fone\ do not
distinguish between significant and insignificant span
differences. This can be misleading and sometimes
surfaces
in surprising ways. Example~4 in Table \ref{tab:examples} shows a reference/candidate pair receiving a relatively high \fone\ score of $0.67$ -- with a completely wrong candidate answer.
\paragraph{Assessment of numbers and units.}
Answers can be equivalent when they express identical values in different units
(and the question does not specify a specific unit), e.g.\ Example~8 in Table~\ref{tab:examples}. 
Similar and frequent problems arise from approximate quantities (e.g.\ the population of a country), metric vs.\ imperial units, percentages and absolute values, and spelled out numbers.

\begin{table*}[h!t]
\centering
\resizebox{\linewidth}{!}{
\begin{small} %
\begin{tabular}{lp{0.15\linewidth}p{0.4\linewidth}p{0.10\linewidth}p{0.12\linewidth}r}

\toprule
\# & Question & Context / Remark & Reference & Candidate & $F_1$ \\
\midrule
1 &  Whose army liberated Warsaw in 1806? & \ldots Liberated by Napoleon's army in 1806, Warsaw was made the capital of the newly created Duchy of Warsaw. \ldots" & Napoleon's & Napoleon & 0.0\\
 & \multicolumn{5}{r}{\textbf{\textit{$	\hookrightarrow$ Reference and candidate are equivalent but \fone~underestimates quality due to tokenization.}}} \\
   \cmidrule(lr){1-6} 
2 & Did Tesla graduate from the university? & In 1875, Tesla enrolled at Austrian Polytechnic in Graz \ldots Tesla claimed that he worked from 3 a.m. to 11 p.m., \emph{no} Sundays or holidays excepted. \ldots Tesla was unprepared and asked for an extension to study, but was denied. He never graduated from the university and did not receive grades for the last semester. & no\footnotemark & He never graduated from the university & 0.0 \\
  & \multicolumn{5}{r}{\textbf{\textit{$	\hookrightarrow$ Equivalence between candidate and reference is easy to establish for humans but fails for automatic evaluation.}}} \\
  \cmidrule(lr){1-6} 
3 & Why is Warsaw's flora very rich in species? & \ldots The species richness is mainly due to the location of Warsaw within the border region of several big floral regions comprising substantial proportions of close-to-wilderness areas (natural forests, wetlands along the Vistula) as well as arable land, meadows and forests. & location & the location of Warsaw within the border region of several big floral regions & 0.14 \\
 & \multicolumn{5}{r}{\textbf{\textit{$	\hookrightarrow$ The candidate adds relevant detail to the reference.}}} \\
  \cmidrule(lr){1-6} 
4 & Other than many sunny days, what characteristic is typical for the weather in Southern California? & Southern California contains a Mediterranean climate, with infrequent rain and many sunny days\ldots & infrequent rain & rain & 0.67 \\
 & \multicolumn{5}{r}{\textbf{\textit{$	\hookrightarrow$ The candidate drops important information.}}} \\
  \cmidrule(lr){1-6} 
5 & What is commonly believed to be the relationship between NP and co-NP? &\ldots It is believed that NP is not equal to co-NP; however, it has not yet been proven. It has been shown that if these two complexity classes are not equal then P is not equal to NP. & NP is not equal to co-NP & P is not equal to NP & 0.84 \\
  & \multicolumn{5}{r}{\textbf{\textit{$	\hookrightarrow$ Superficially high token overlap, yet different and wrong answer.}}} \\
  \cmidrule(lr){1-6} 
6 & What types of teachers are retiring the most? & \ldots Excellent job opportunities are expected as retirements, especially among secondary school teachers, outweigh slowing enrollment growth; \ldots & secondary school teachers & secondary school & 0.8  \\
  & \multicolumn{5}{r}{\textbf{\textit{$	\hookrightarrow$ Candidate answer is only equivalent given the question.}}} \\
  \cmidrule(lr){1-6} 
7 & Who did Tesla think would run the world of the future? & In 1926, Tesla commented on the ills of the social subservience of women and the struggle of women toward gender equality, and indicated that humanity's future would be run by ‘Queen Bees.’ & women & Queen Bees & 0.0 \\
  & \multicolumn{5}{r}{\textbf{\textit{$	\hookrightarrow$ Candidate answer is only equivalent given the context.}}} \\
  \cmidrule(lr){1-6} 
8 & How much do researchers now think sea levels will rise from 1990 to 2100? & \ldots When the researchers' analysis was applied to the possible scenarios outlined by the Intergovernmental Panel on Climate Change (IPCC), the researchers found that in 2100 sea levels would be 0.5–1.4 m [50–140 cm] above 1990 levels. \ldots & 50–140 cm & 0.5–1.4 m & 0.0\\
  & \multicolumn{5}{r}{\textbf{\textit{$	\hookrightarrow$ Candidate is identical to the reference but the metric is not sensitive to units.}}} \\
\bottomrule
\end{tabular}
\end{small}
}
\vspace{-1.5ex}
\caption{Examples where $F_1$ score does not adequately represent the quality of the candidate answer.}
\label{tab:examples}
\vspace{-1ex}
\end{table*}

\subsection{Further limitations}

We find \fone\ has several more specific shortcomings that make it less suitable for QA evaluation.
\vspace{-1.5ex}
\paragraph{Asymmetry.}
A candidate answer can improve over the reference by adding
relevant information, such as Example 3 in Table \ref{tab:examples}. In that case it should get credit in evaluation, even
though it is not strictly equivalent. Conversely,
omitting relevant information (which is present in the reference answer)
in the candidate answer should be discouraged.

On the other hand, if the candidate answer removes irrelevant or misleading
information, it also improves the reference answer, and, even though it is
not strictly equivalent, should get full credit in evaluation. 
Both EM and \fone\ fail to recognize such cases, because they are symmetric.
\vspace{-1.5ex}
\paragraph{The question matters.}
Whether two answers are equivalent might depend on the question. In Example~6 in Table \ref{tab:examples} `secondary school' is equivalent to `secondary school teachers' only because the question explicitly asked for teachers.

Note that this is a particular example of a \emph{fused-head construction}, which occur frequently in machine reading datasets. \citet{yanai-2019-fused-head} discuss this phenomenon
for numerical cases.
\vspace{-1.5ex}
\paragraph{The context matters.}

More rarely than with questions, the context can determine
whether two answers are equivalent. Consider Example~7 in Table \ref{tab:examples}. `Queen Bees' only qualifies as a possible match for the reference answer `women', because it is used as a metaphor in the context.

\footnotetext{Note that this annotation violates the SQuAD guidelines. However, human raters are easily able to establish equivalence of these answers.}

%% file: sections/30-data.tex
\section{The AE task definition and dataset}
\label{sec:data}

To address the issues mentioned in Section~\ref{sec:issues}, we require
the answer equivalence relation to be \emph{asymmetric and to
be conditional on both question and context}. We want an
ideal metric to give credit for a candidate answer that is at least as good as the reference, i.e. it should capture all the important information in the reference and not add irrelevant, or worse, misleading information. A candidate answer that improves
over the reference by either removing misleading or irrelevant information, or adding
more relevant information, should receive full credit. More formally, we define:
\begin{definition}[Answer Relation]
Let $q$ be a query and let $a_1, a_2$ be answers contained in a context $c$. Then \emph{$a_2$ is a good answer in place of $a_1$} if both the following are satisfied:
\begin{itemize}[leftmargin=*, noitemsep]
    \vspace{-4pt}
    \item[(i)] $a_2$ does contain at least the same (or more) relevant information as $a_1$,
               taking into account $q$ and $c$; in particular it does not omit any relevant information present in $a_1$.
    \item[(ii)] $a_2$ contains neither misleading or excessive superfluous information not present in $a_1$,
                taking into account $q$ and $c$.
\end{itemize}
\end{definition}

Note that this approach does not aim to replace the regular QA annotations, but expands on them to create larger, more inclusive sets of acceptable answers. Most studies rely on token-level measures to approximate this expansion. However, as detailed in Section~\ref{sec:issues}, such measures are inadequate.

\subsection{Rating task}
\label{sec:questions}

We design the rating task for answer equivalence as follows: the raters are presented with (i) the question, (ii) context from Wikipedia that contains the answer text, (iii) the reference answer (referred to as `first answer'), and (iv) a candidate answer (referred to as `second answer'). They are then asked the following yes/no questions in sequence:

\begin{footnotesize}
\begin{enumerate}%
    \item[Q1] Is the second answer a completely different answer?
    \item[Q2] Would using the second answer in place of the first answer convey at least the same information to someone asking this question, without leaving out any important information nor adding any misleading or superfluous information? \\
    (Note that either adding important information or removing superfluous or misleading information in the second answer from the first answer would still convey at least the same information.)\footnote{In initial rounds we asked a much shorter version of this question: `Can the second answer be used in place of the first answer?' and had an exact definition of the differences we were looking for in the overall task description. As there was frequent confusion on the details we noticed that we had more success giving the precise wording here. Note that we deliberately don't use the notion of \emph{equivalence} in our task to avoid varying interpretations.}
    \item[Q3] Does the second answer remove important information?
    \item[Q4] Does the second answer add misleading or excessive superfluous information?
\end{enumerate}
\end{footnotesize}

If a rater answers `yes' to the first question, the rating is terminated and the following questions are not shown. Similarly, if a rater answers `yes' to the second question, the rating ends and $Q3$/$Q4$ are not shown. Otherwise, all four questions are answered.

\subsection{Data collection}

We use the above task to annotate examples generated from the \squad\ dataset~\cite{rajpurkar-16}, labelling examples from both \emph{train} and \emph{dev}.

For the training examples, we partition the \squad\ train set 5-ways at random, and train five Albert models~\cite{albert-paper-19}, each excluding one of the partitions (i.e. training on 80\% of the available data). We then use each model to generate predictions for the unseen examples in its excluded partition, thereby generating predictions for the entire \squad\ train set. We rate all examples where the prediction does not match the reference.

Three different models are used to make prediction on the \squad\ dev set:   BiDAF~\cite{bidaf-paper-16}, XLNet~\cite{yang-xlnet}, and Luke~\cite{luke-paper-16}. 
As before we remove predictions that match any of the (up to $6$) reference answers. Otherwise we pair the prediction with reference and annotate all combinations.

For XLNet predictions we collect up to 4 annotations\footnote{We used XLNet for quality control and to estimate rater agreement; see below.}; for Luke and BiDAF we obtain only a single annotation. Overall, we collect $14,170$ annotations for $8,565$ (question, context, reference, candidate)-tuples for $4,369$ non-EM predictions. See the Appendix (Table \ref{tab:datastats}) for detailed statistics.

\subsection{Train / Dev / Test splits}
We provide two partitionings of the dev data: either split by the system producing the candidate as introduced above, see Table \ref{tab:datastats}, or in a 30/70 dev/test split, see Table~\ref{tab:datastats_dev_test}. For the latter we select examples with no document overlap between dev and test sets. In either case, the training data is the same.

\begin{table}[t]
\centering
\resizebox{\linewidth}{!}{
\begin{tabular}{lrrrr}
\toprule
Count & AE Train & AE Dev & AE Test & Total \\
\midrule
AE-examples & 9,090 & 2,734 & 5,831 & 17,655\\
Ratings & 9,090 & 4,446 & 9,724 & 23,260\\
\bottomrule
\end{tabular}
}
\vspace{-1ex}
\caption{Data statistics for human rated (question, context, reference, candidate) 4-tuples.}
\label{tab:datastats_dev_test}
\vspace{-0.7\baselineskip}
\end{table}

\subsection{Answer Equivalence definition}
Each annotation consists of four binary answers to the questions given in Section \ref{sec:questions}. We define the candidate to be \emph{equivalent} to the reference, if it is rated as (1) not completely different (i.e. $Q1$ is answered `no') and (2) containing at least as much relevant information  and not more irrelevant information as the gold answer (i.e. $Q2$ is answered `yes'). In all other cases we define the candidate to be \emph{not equivalent} to the reference.

\subsection{Quantitative analysis}

Overall our raters consider $55\%$ of the pairs that are not exact matches to be equivalent. At 69.9\% that rate is higher for the candidates from \squad\ train but also differs across systems. As expected and shown in Table \ref{tab:equivalence_rate}, ratings for AE examples produced by BiDAF are less likely to be rated equivalent than those produced by XLNet or Luke.

We collect a total of $14,170$ ratings on \squad dev, for $8,565$ AE-examples, i.e. (context, question, reference, candidate)-tuples (Table \ref{tab:datastats} in the Appendix). $6,062$ examples have a single rating, $17$ have two, $1,870$ have three and $616$ have four. We aggregate the ratings per example using majority voting.\footnote{For the very few ($<25$) ties, we take a random decision.} Among all examples with multiple annotations, over $88\%$ have full agreement between raters. Similarly, selecting a random pair of ratings for the same example, has a $92\%$ chance of agreement. Finally, we compute Krippendorff's $\alpha$ at $0.84$, confirming good inter-annotator agreement.\footnote{According to \citet{krippendorff04}, it is ``customary to require $\alpha > 0.8$''.}

\begin{figure}
    \centering
    \includegraphics[width=0.98\linewidth]{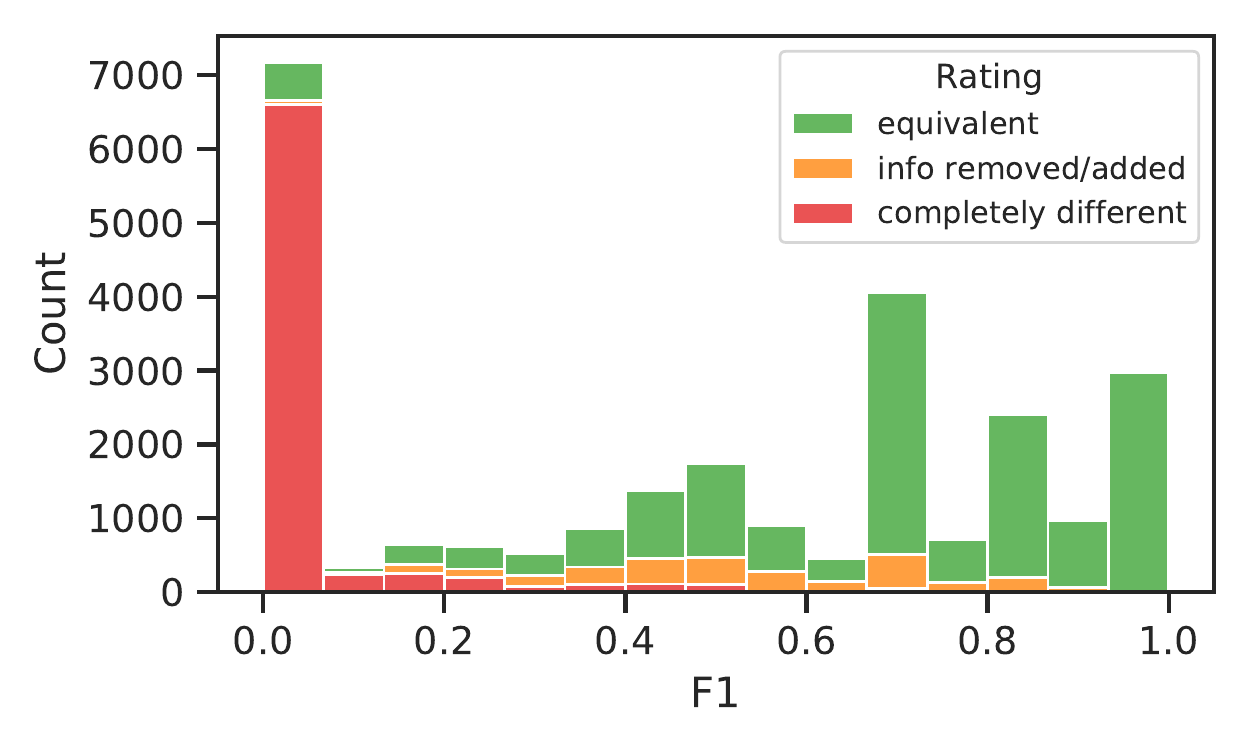}
    \vspace{-1\baselineskip}
    \caption{Histogram of $F_1$ scores, colored by their manually annotated equivalence ratings. In this figure we differentiate between two classes of non-equivalent answers, those that are completely different and those that either remove relevant information or add distracting or misleading information.}
    \label{fig:f1_vs_rating}
    \vspace{-1\baselineskip}
\end{figure}

As expected, Figure \ref{fig:f1_vs_rating} shows the vast majority of \emph{completely different answers}
are cases of no token overlap between reference and candidate answer. Interestingly, all buckets with $\fone>0$ contain a sizeable portion of equivalent answers.

\subsection{Qualitative analysis}

Comparing raters' judgment on the Answer Equivalence task with F1 scores shows typical disadvantages, and limitations of F1. In the area of high F1 scores (i.e.\ $>0.6$), raters find answers with significantly changed meaning, despite having substantial overlap with the reference answer. For example, \emph{`O notation'} can not necessarily be understood as \emph{`Big O notation'} ($F1=0.8$), or \emph{`June'} is not equivalent to \emph{`every June'} ($F1=0.66$).

On the other hand, in the area of low F1 scores (e.g. $<0.3$), we can see answers that convey the same core content as the reference but with added information that an informed human would likely give.
For example, for a question \emph{`In the most basic sense what did a Turing machine emulate?'}, an answer \emph{`a very robust and flexible simplification of a computer'} is more informative than the bare minimum reference \emph{`a computer'}.

Moreover, the candidate can add crucial information, thereby improving over the reference: e.g. for the question: \emph{`What acquired condition results in immunodeficiency in humans?'}, raters rate positively an answer \emph{`HIV/AIDS, or the use of immunosuppressive medication'}, despite the reference being only \emph{`HIV/AIDS'}.

\subsection{Quantifying limitations of \fone}

In our data we observe three further issues with \fone.
Firstly, having values on a scale from $0$ to $1$ gives the impression
of a gradual rating, with e.g. $0.6$ being \emph{twice as correct}
as $0.3$. Almost all of the answers with a rating greater than $0$
were rated as \emph{equivalent}.

Secondly, we observe that \fone\ systematically underestimates
model performance compared to human judgment. Table~\ref{tab:metric_comparison} shows that both EM and \fone\ severely underestimate the quality of predictions.

Lastly, \fone\ is highly dependant on the number of available
references. Table~\ref{tab:references_comparison} quantifies how
\fone\ becomes a better estimator of human judgment when
more references become available. This is an issue because
additional references are not available or expensive and laborious to generate.

%% file: sections/40-model.tex
\section{Predicting Answer Equivalence}\label{sec:predicting_AE}

In the previous section we defined the AE task and discussed the value of such annotations. In practice, however, obtaining human ratings is slow and expensive. This is especially true for open-domain or generative QA models, where the space of possible answers is infinite. Instead, we train a classifier to predict AE for any answer pair.

We find the classifier's score to be more accurate (\S\ref{sec:measures_comp}), and provide practical gains (\S\ref{sec:conformal_main}). We conjecture that the high performance of the classifier is due to the easier nature of the AE task compared to QA. While, under some assumptions, the question-answering problem is NP-complete~\citep{weston2015aicomplete}, the verification of a candidate answer against an already given true answer is significantly simpler.
\paragraph{The classification task.}
Given the four-tuple (context, question, reference answer, candidate answer), predict the equivalence of the answers.

\subsection{Bert Match (\ourscore)}
To achieve the approximate AE predictions, we train a BERT-Base~\cite{devlin-19} model on the AE training set. (Training details can be found in Appendix~\ref{app:bert}.)

\subsubsection{Comparing input variations} \label{sec:input_variations}

We experiment with three different settings for incorporating  the question and context:

\begin{itemize}[leftmargin=*]
    \item[(i)] \textbf{Only answers:} Just using the two answers as model input; more precisely: \emph{[CLS] candidate [SEP] reference [SEP]}
    \item[(ii)] \textbf{Question and answers:} We present the question and the two answers; i.e. \emph{[CLS] candidate [SEP] reference [SEP] question [SEP]}
    \item[(iii)] \textbf{Context, question and answers:} We present all available information: \emph{[CLS] candidate [SEP] reference [SEP] question [SEP] context [SEP]}
\end{itemize}

\begin{table}[t]
\begin{center}
\begin{tabular}{lr}
\toprule
Mode & Accuracy \\
\midrule
Only answers & 89.27 \\
Question and answers & \textbf{90.70} \\
Context, question and answers & 89.53 \\
\bottomrule
\end{tabular}
\end{center}
\vspace{-0.5\baselineskip}
\caption{Comparison of classification accuracy between
best model performance, using as input different parts
of the available information: (i) just answers, (ii) question
and answers, (iii) context, question and answers. The
accuracy is reported on XLNet predictions on the SQuAD
dev set.
}
\label{tab:modes}
\vspace{-0.5\baselineskip}
\end{table}

The classification accuracy results comparing the three different models are shown in Table~\ref{tab:modes}. A model that just sees both answers (i)
already performs well on the task, improving over other metrics (Table~\ref{tab:dev_accuracy}). 
Adding the question (ii) further improves the classification accuracy, possibly due to the need of disambiguating the relevance of the answer. For example, in line 6 of Table~\ref{tab:examples} both answers with or without the word ``teachers'' are equivalent since the context is clear from the question. If the question was about schools the two answers would not be equivalent.

Finally, adding the context (iii) degrades the performance compared to (ii). We hypothesize that this is for three reasons: First, the number of training examples where the context contains pertinent information is
insufficient to be used productively in our dataset.\footnote{Only around 2-3\% of examples require a closer assessment of the context to make an accurate AE assessment, measured by (the authors) inspecting random $100$ examples.} Second, cases that require assessment of the context are harder examples, both for the model, and
for humans, which may lead to less consistent annotations and add noise during the learning process.
Third, the context is the longest part of the input and may contain many irrelevant parts. 
We leave further exploration to future research and
use variant (ii) in the following.

\subsubsection{Accuracy of \ourscore predictions} \label{sec:measures_comp}

Table~\ref{tab:dev_test_performance} shows that our \ourscore model achieves high accuracy and  correlation on the task of predicting the human equivalence
ratings, significantly improving over the baselines.
The gain in accuracy is consistent across the choice of QA system as shown in Table~\ref{tab:dev_accuracy}. We include BERTScore \citep{bertscore}, Bleurt \citet{sellam-etal-2020-learning}, and LERC \citet{chen-et-al-2020} as additional baselines.\footnote{BERTScore uses the published uncased BERT-Base model.}

\begin{table}[ht]
\begin{center}
\resizebox{0.48\textwidth}{!}{%
\begin{tabular}{lrrrr}
\toprule
\multirow{2}{*}{Metric (\%)} &
  \multicolumn{2}{c}{AE Dev} &
  \multicolumn{2}{c}{AE Test} \\
\cmidrule(lr){2-3} \cmidrule(lr){4-5}
& Acc & $\rho$ & Acc & $\rho$ \\
\midrule
EM    & 54.00 & -- & 54.69 & -- \\
\fone & 75.57 & 72.43 & 75.95 & 69.07 \\
\fone (tuned) & 84.39 & 72.43 & 82.93 & 69.07 \\
BertScore & 73.55 & 57.08 & 70.27 & 52.40 \\
BertScore (tuned) & 73.80 & 57.08 & 70.87 & 52.40 \\
Bleurt 2.0 & 72.02 & 66.59 & 73.02 & 63.65 \\
LERC (tuned) & 81.96 & 71.27 & 80.74 & 67.81 \\
\midrule
\ourscore & \textbf{89.38} & \textbf{79.92} & \textbf{89.66} & \textbf{79.09} \\
\ourscore (tuned) & \textbf{89.99} & \textbf{79.92} & \textbf{89.84} & \textbf{79.09} \\
\ourscore (symmetrized) & 86.95 & 72.27 & 84.56 & 71.57 \\
\bottomrule
\end{tabular}
}%
\end{center}
\caption{BEM classification accuracy and correlation (Spearman's $\rho$) with human judgment. For the tuned version we select an optimal threshold on the train set. }
\label{tab:dev_test_performance}
\end{table}

To compute accuracy we either threshold at $0.5$ or tune a threshold such that accuracy on the train set is optimal.

\begin{table}[ht]
\begin{center}
\begin{tabular}{lrrr}
\toprule
Metric & \multicolumn{3}{c}{Accuracy, predictions from} \\
\cmidrule(lr){2-4} &
BiDAF & XLNet & Luke \\
\midrule
EM & 61.23 & 44.26 & 41.07 \\
\fone & 79.26 & 77.33 & 74.64 \\
BERTScore & 69.06 & 74.47 & 73.83 \\
\midrule
\ourscore & \textbf{88.83} & \textbf{90.70} & \textbf{89.48} \\
\bottomrule
\end{tabular}
\end{center}
\vspace{-0.5\baselineskip}
\caption{AE classification accuracy on predictions from different models.
Existing metrics fail to capture the equivalence of answers, and significantly under-perform compared to \ourscore. 
}
\label{tab:dev_accuracy}
\end{table}

\subsection{AE for QA performance evaluation} \label{sec:ae_perf_eval}

As previously discussed, accurately assessing the performance of QA models is a significant challenge. Even in the presence of multiple gold answers and predefined normalization rules, models might output correct answers outside of the annotated answer bank.
Thus EM provides a pessimistic lower bound on the true accuracy, and AE metrics can provide a more realistic assessment.

\begin{table}[h]
\begin{center}
\begin{tabular}{lrrr}
\toprule
Metric    & \multicolumn{1}{c}{BiDAF}  & \multicolumn{1}{c}{XLNet} & \multicolumn{1}{c}{Luke} \\
\midrule
EM        & $71.62_{ \pm 0.44}$  & $88.60_{ \pm 0.31}$  & $89.76_{ \pm 0.30}$  \\
\fone     & $80.79_{  \pm 0.34}$ & $94.15_{  \pm 0.19}$ & $94.99_{  \pm 0.18}$ \\
\ourscore & $84.60_{  \pm 0.35}$ & $96.00_{  \pm 0.19}$ & $97.03_{  \pm 0.16}$ \\
Human     & $83.25_{  \pm 0.36}$ & $96.67_{  \pm 0.18}$ & $96.76_{  \pm 0.17}$ \\
\bottomrule
\end{tabular}
\end{center}
\vspace{-0.5\baselineskip}
\caption{Comparison of metrics on full dev set
predictions from Luke, XLNet and BiDAF (at the time of writing their SQuAD 1.1 leaderboard positions are 1, 2 and 52).
Confidence intervals obtained with bootstrapping.
}
\label{tab:metric_comparison}
\end{table}

In order to evaluate the validity of the approximate metrics, we compute the accuracy of three representative models according to each equivalence metric against the \squad gold answer set. In addition, we use AE annotations to measure the true model performance.
As Table~\ref{tab:metric_comparison} shows, indeed the EM is significantly lower than the true accuracy. \fone provides a slightly more optimistic evaluation, though still below the true score in about 2-3 points. The learned \ourscore-based accuracy computation is closest to the true performance of the model.

Next we show that the learned metric is much more robust
with respect to the number of available references (Table~\ref{tab:references_comparison}).
At the time
of writing, Luke (rank 1) outperforms XLNet (rank 2) on the the SQuAD 1.1 leaderboard. However,
if only one gold reference was available, a higher degree
of accuracy would be needed to distinguish them with \fone compared to the learned metric. 
Even with just a single reference, \ourscore is
much closer to the human judgment shown in Table~\ref{tab:metric_comparison}.

\begin{table}[t]
\begin{center}
\resizebox{0.48\textwidth}{!}{%
\begin{tabular}{lcccc}
\toprule
Metric &
\multicolumn{2}{c}{$F_1$} &
\multicolumn{2}{c}{\ourscore} \\
\cmidrule(lr){2-3} \cmidrule(lr){4-5}
\# References & 1 & all & 1 & all \\
\midrule
BiDAF & 73.31 & 80.79 (+7.48) & 80.43 & 84.60 (+4.17) \\
XLNet & 86.63 & 94.15 (+7.52) & 92.13 & 96.00 (+3.87) \\
Luke  & 87.30 & 94.99 ( +7.69) & 92.98 & 97.03 (+4.05) \\
\bottomrule
\end{tabular}
}%
\end{center}

\vspace{-0.5\baselineskip}
\caption{Comparison of metrics with respect to the number of
references on dev set
predictions from Luke, XLNet and BiDAF.
The parenthesis show the difference
between using a single reference and using all available references (up to 6, on average 3).
}
\label{tab:references_comparison}
\end{table}

To assess how well \ourscore generalizes to more challenging tasks, we run a baseline system on NQ-Open \citep{lee-etal-2019-latent} consisting of a BM25-based retriever and a BERT-based extractive reader. We sample 300 predictions from the dev set, ignoring exact matches. In 87\% of these examples we find our manual independent assessment to agree with \ourscore. 
This is evidence, that \ourscore does not just generalize well from Albert
answers to other models on SQuAD, but also to other more difficult datasets
and QA settings such as open-domain.

%% file: sections/41-conformal.tex
\section{Example application: Returning small and accurate prediction sets} \label{sec:conformal_main}

Beyond the value of accurate performance assessment, identifying when the model answers are right can provide immediate practical gains. We demonstrate this on the common setting of constructing \emph{prediction sets} (e.g., the top-$k$ model's predictions). Here, the accuracy is measured by whether a correct answer is included in the predicted set. The size of the set ($k$) is typically calibrated on an evaluation set, to the desired  accuracy. Reliably identifying high-ranking correct answers allows us to effectively reduce the size of the sets (smaller $k$) while keeping them accurate.

Here, we use the \emph{conformal prediction} (CP) framework ~\citep{cp_tutorial2, cp_tutorial} to construct provably accurate prediction sets. Unlike top-$k$, CP set size is dynamically determined per input with instance-wise hypothesis testing to marginally satisfy the user-specified target accuracy. 
Specifically, given an exchangeable calibration set of pairs of questions and correct answers, we measure their nonconformity scores (here, the negative predicted probability). Then, for a new question, CP returns a set of candidate answers by including all answers with nonconformity scores smaller than the inflated $\alpha$ Quantile of the calibration scores, where $\alpha$ is the target accuracy.
\looseness=-1

While CP was originally defined for a single label per input, \citet{fisch2021efficient} recently presented \emph{expanded admission} CP to support multiple answers. This extension allows  leveraging an answer equivalence function to reduce the size of the CP sets while preserving the accuracy guarantee.

\begin{table}[t]
    \centering
    \resizebox{0.48\textwidth}{!}{%
    \begin{tabular}{l|ccccc}
        \toprule
        Admission& \multicolumn{5}{c}{Target accuracy} \\
        function & 99\% & 95\% & 90\% & 80\% & 70\% \\
        \midrule
        \multicolumn{6}{l}{\emph{Exact admission (human annotations):}}  \\
         SQuAD labels & 20.00 & 18.16 & 11.31 & 5.31 & 3.28 \\
         + AE labels & \textbf{17.29} & \textbf{8.41} & \textbf{4.31} & \textbf{2.02} & \textbf{1.37} \\
         \midrule
         \multicolumn{6}{l}{\emph{Approximate admission (by model's AE predictions):}}  \\
         \fone & 20.00 & 13.92 & 6.86 & 2.97 & 1.99 \\
         \ourscore & \textbf{17.69} & \textbf{8.74} & \textbf{4.26} & \textbf{2.16} & \textbf{1.51} \\
         \bottomrule
    \end{tabular}
    }%
    \caption{Average size of conformal prediction set (lower is better) per target accuracy (i.e., ratio of sets that include a correct answer). All methods empirically meet the target accuracy. The AE examples allow identifying higher ranked correct answers for calibrating the decision threshold, thereby producing smaller prediction sets that are still accurate. In the absence of AE labels for calibration, \ourscore's approximate predictions (post FPR correction) are similarly effective.}
    \label{tab:conformal_results}
\end{table}

\begin{figure}[ht]
     \centering
     \begin{subfigure}[b]{0.45\textwidth}
         \centering
         \includegraphics[width=\textwidth]{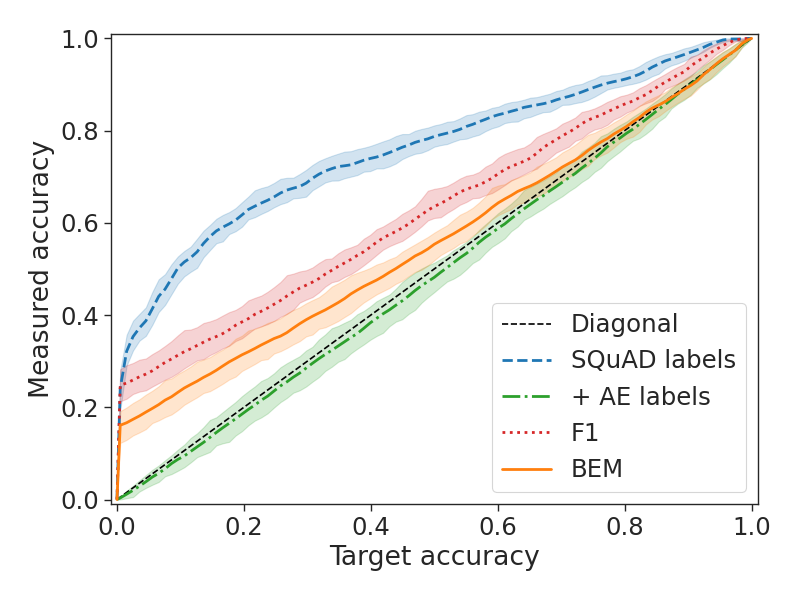}
         \caption{Measured vs.\ target accuracy of the prediction sets.}
     \end{subfigure}
     \begin{subfigure}[b]{0.45\textwidth}
         \centering
         \includegraphics[width=\textwidth]{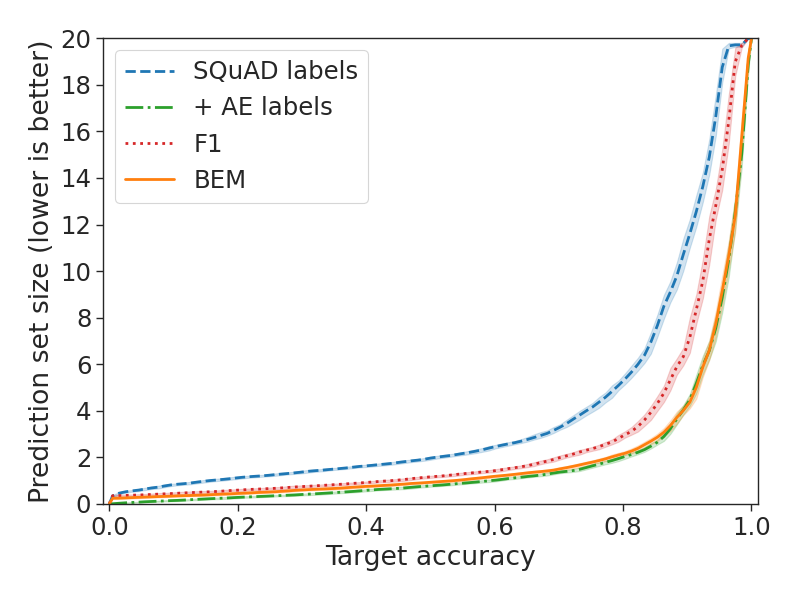}
         \caption{prediction sets size per target accuracy (lower is better).}
     \end{subfigure}
        \caption{Results of the conformal prediction sets per user-defined target accuracy using different equivalence measures for calibration. Note that \squad and AE labels are exact measures while \fone and \ourscore are approximated (and therefore require a statistical correction before calibration). Yet, \ourscore provides similar efficiency gains as the AE labels (i.e., small prediction sets) while meeting the desired target accuracy or higher. Exact values for reference target accuracy are given at Table~\ref{tab:conformal_results}.}
        \label{fig:conformal}
\end{figure}

We follow the setting of \citet{fisch2021efficient} and use the collected AE labels to construct accurate CP sets for SQuAD 1.1 with Luke's scores. As Table~\ref{tab:conformal_results} shows, this leads to significantly smaller sets, sometimes less than half the size. For example, to achieve 90\% accuracy, calibrating with SQuAD labels results in an average of 11.31 answers per question. Expanding the admission to include AE labels reduces this size to only 4.31 answers. 

\citet{fisch2021efficient} also discuss the use of an approximate admission function (see their appendix B). This setting is useful for when high-quality annotations for calibration are missing. We evaluate this here and experiment with using either \fone or \ourscore AE predictions. As Table~\ref{tab:conformal_results} indicates, we find \ourscore to be as effective as the AE evaluation labels for reducing the number of required answers per target accuracy level. 

Figure~\ref{fig:conformal} shows the results over the full range of target accuracy ($\alpha$) averaged over 50 trials, with 16 and 84th percentiles shown in shaded color.
Additional details are available in Appendix~\ref{app:conf}.

%% file: sections/69-related.tex
\section{Related Work}

\paragraph{Answer equivalence.}

Most similar to our work, \citet{risch2021semantic} annotate 1k answer pairs from the SQuAD gold answers with a similarity score and train a model that classifies the concatenated strings. They focus on the symmetric string-similarity problem whereas we aim for asymmetric equivalence conditioned on the answer.

Another similar effort is the MOCHA dataset \citet{chen-et-al-2020}. The authors also collect annotations on answer candidates and train a learned metric. However, they focus on generative question answering. Moreover, their methods for collecting
candidate answers, the selection of datasets and the rating task differ
considerably from our work.

\citet{chen-etal-2021-pw} use natural language inference to verify
predictions from QA systems by converting question and answer into a statement. They use this to improve predictions in a
setting where no gold answer is known, but
one could potentially employ similar methods to compare a predicted
answer to a gold answer.
\paragraph{Text similarity.}

To our knowledge, \citet{breck-1999-aw} first used Token \fone\ for automatic
evaluation of their ``Sys called Qanda''.
\citet{chen-etal-2019-evaluating} find that $F_1$ is a reasonable metric for extractive QA tasks but fails to account for the higher variability of surface form in generative QA.

Besides Token \fone\ and Exact Match, other popular metrics for text 
comparison have been tried for question answering, but are not
widely in use: BLEU \cite{bleu}, Rouge \cite{lin-2004-rouge} and METEOR \cite{banerjee-lavie-2005-meteor}.

\citet{yang-etal-2018-adaptations} identify the need for methods that go beyond lexical overlap, trying to adapt ROUGE and BLEU to better fit answer comparison, but focusing on just ``yes-no'' and ``entity'' questions.
Using a different approach, \citet{Si:Zhao:Boyd-Graber-2021} propose to expand entities in gold answers with aliases from Freebase \cite{freebase08} to improve exact match reliability.

A good overview of different string distance metrics, in the context of name-matching tasks, can be found in \cite{Cohen2003ACO}. %
Metrics based on the Wasserstein distance have also been proposed \cite{kusner-etal-2015-ls, clark-etal-2019}.

In addition to automatic metrics, the NeurIPS 2020 EfficientQA Competition \citep{min-et-al-2021} uses manual annotations to reward correct answers not contained in the gold answers. In contrast, we focus on only rewarding answers that are equivalent to one of the gold answers.

An important topic that is tangentially related to our work is the
question of ambiguous questions (cf.\ \citet{min-et-al-2020,min-et-al-2021}). Our approach to answer equivalence
could be useful for detecting the presence of multiple clusters of equivalent
answers, suggesting an ambiguous question.
\paragraph{Learned metrics.}
Recently string based metrics have been replaced by learned metrics for various Natural Langauge Generation (NLG) tasks. Examples include BLEURT \citep{sellam-etal-2020-bleurt} and COMET \citep{rei-etal-2020-comet} for machine translation evaluation, which \citet{kocmi2021ship} find to correlate better with human judgments than e.g. the popular BLEU metric \citep{bleu}.

For question answering evaluation, \citet{chen-etal-2019-evaluating} propose a variant of BERTScore \citep{bertscore} where the answer tokens are contextualized by adding the question and context but can't show improvements over $F_1$ for extractive QA. 

%% file: sections/70-conclusion.tex
\section{Conclusion}

We present a systematic data-driven analysis of the shortcomings of token-level measures, EM and \fone, for evaluating QA systems against a set of reference answers.

We design an answer equivalence (AE) task that directly captures the desired relation between candidate and reference answer. Also, we collect a large number of annotations for both evaluating and training equivalence models, as well as quantitatively assessing the performance of QA systems. 

Beyond relying on human AE annotations for evaluation, we trained a BERT matching (BEM) model and showed that it generalized well to new QA models and evaluation questions. Specifically, BEM allowed a significantly better performance assessment for the QA systems compared to the token-level or other similarity measures. We also demonstrated the value of AE on a practical application beyond QA performance assessment.

We hope releasing our data will contribute to further development
of better metrics and improve the evaluation and usability of QA systems.

%% file: sections/80-ethics.tex
\section*{Ethical Considerations}
As our work involves human participants, we point out that all annotators provided informed consent and no personally identifiable information (\abr{pii}) was collected or will be released. The collected data has been vetted for presence of \abr{pii} as well as offensive language through heuristics and random sampling. 

The annotators received fair compensation with respect to local markets, but said compensation was not tied to speed or accuracy to prevent distorting the motivation. Intrinsic motivation has been shown to produce higher quality results~\cite{payenough}. 

The released data and the experiments we conducted are in
English, therefore we do not claim generalization of our findings across languages. However, we believe that the proposed methods could be
applied in other languages using other available corpora as sources.

%% file: sections/81-limitations.tex
\section*{Limitations}

Our current work has several limitations that we hope will inspire future research and extensions to our framework.

One limitation is our focus on short answers, which is in line with much of the
current QA research. Some datasets (e.g., Natural Questions) have introduced long answers.
Our setting would not be directly applicable there, as equivalence between long answers could be harder to determine. However, BEM could be a building block in a more complex
aggregated comparison (e.g., ``summarizing'' the long answer with a machine reading model and comparing the
``summaries'').

A further limitation, that is discussed in more detail in Section~\ref{sec:input_variations}, is our final BEM is 
not using the context available for determining equivalence. This is an advantage as it allows BEM to be readily extended to settings where context is unavailable. On the other hand, we expect AE on certain QA domains to perform better with context, and encourage future research to explore when context is useful.

The research was done on English language datasets (SQuAD, Natural Questions), but the same methodology
for data collection and model training should extend to other languages given the existance of high-quality
QA datasets for the languages in question. A further limitation may be multi-lingual answers. Translation alone may
not be enough to establish equivalence, given potential subtle differences in semantics between languages.

Also, our main experiments focus on machine reading models. In Section~\ref{sec:ae_perf_eval} we discuss extensions to NQ-Open and show promising initial results.

A more specific challenge regarding Open QA is the use of generative models for QA. Here the generative
model may add more specific details (e.g., a middle name, or a more precise date) that is not supported
by the available context and may
potentially include false hallucinations. Since our model has been trained only with extracted answers, we saw in a small
experiment that it will generally accept these more specific answers as equivalent answers. Note that this
is also a more general challenge when annotating answers for QA datasets and not only for answer equivalence---possibly requiring specific background knowledge and/or extra time/resources when verifying an answer.

Our approach does not directly address the (potential) temporal or spatial dimension of questions/answers.
While our definition allows us to handle answers that vary by time/location, in rare cases the equivalence
of an answer may depend on the time/location it is given. For example, the two answers ``February 2022'' and
``4 months ago'' to the question ``When were the last Olympics?'' would only be equivalent in June 2022.
This is a limitation of the QA setting, and could be solved by recording the exact date/time and location
of the annotation and adding it to the context.

Lastly, some more specific question answering tasks are out of scope for us and left for future work. For example, conversational
question answering (e.g., CoQA) may need different processing of questions and context.

%% file: sections/90-appendix.tex
\begin{table}
\begin{center}
\begin{tabular}{lrrr}
\toprule
Metric & Threshold & AE Dev & AE Test \\
\midrule
\fone & 0.2 & 85.06 & 82.47 \\
LERC & 2.52 & 81.97 & 80.74 \\
\midrule
\ourscore & 0.56 & \textbf{89.99} & \textbf{89.80} \\
\bottomrule
\end{tabular}
\end{center}
\vspace{-0.5\baselineskip}
\caption{BEM classification accuracy with optimal threshold tuned on AE train set.}
\label{tab:dev_test_performance_threshold}
\end{table}

%% file: sections/98-data-stats.tex
\section{Additional dataset statistics}

We provide additional statistics about the collected annotations in Table~\ref{tab:datastats}
and Table~\ref{tab:equivalence_rate}.

\begin{table}[t]
\centering
\resizebox{\linewidth}{!}{
\begin{tabular}{lrrrr}
\toprule
Count & XLNet & BiDAF & Luke & Total \\
\midrule
Non-exact matches & 1,205 & 3,000 & 1,082 & 4,369\\
AE-examples & 2,448 & 5,655 & 2,240 & 8,565\\
Ratings & 7,932 & 7,522 & 4,590 & 14,170\\
\bottomrule
\end{tabular}
}
\vspace{-1ex}
\caption{Data statistics for human rated (question, context, reference, candidate) 4-tuples. Since several systems might produce the same non-matching candidate, numbers in the Total column are lower than row sum.}
\label{tab:datastats}
\end{table}

\begin{table}[t]
\centering
\resizebox{\linewidth}{!}{
\begin{tabular}{lSSS}
\toprule
\% & {BiDAF} & {XLNet} & {Luke} \\
\midrule

Equivalent candidates & 40.36666666666667 & 66.05809128630705 & 66.728280961183\\
Equivalent AE-examples & 32.30769230769231 & 55.9640522875817  & 53.66071428571428\\
Equivalent ratings & 38.76628556235044 & 55.73625819465456 & 58.93246187363834\\
\bottomrule
\end{tabular}
}
\vspace{-1ex}
\caption{Percentages of equivalent answer candidates according to human ratings. Note that this only takes answer candidates into account that did not match any of the reference answers.}
\label{tab:equivalence_rate}
\vspace{-\baselineskip}
\end{table}

%% file: sections/99-a-bert-details.tex
\section{BEM training details}
\label{app:bert}

For BEM training we finetune the published BERT-Base (uncased, 12-layer, 768-hidden, 12-heads, 110M parameters) checkpoint on the training examples for one epoch, using a JAX-based BERT implementation. We use a batch size of 64 and a learning rate of 1e-4 with the Adam optimizer. We did not perform a search for optimal hyperparameters. The training on a TPU v2 takes less than 5 minutes.

%% file: sections/99-b-conf-details.tex
\section{Conformal prediction sets experimentation details}\label{app:conf}
The experiments in section~\ref{sec:conformal_main} use the Conformal Prediction (CP) framework~\citep{angelopoulos2021gentle, cp_tutorial2, vovk_random}. While recent work found CP to be useful in many practical applications such as  medical image segmentation~\citep{bates2021distribution} and adaptive computation Transformers~\citep{schuster-etal-2021-consistent, Schuster2022CALM}, one of CP challenges is in reducing the size of the prediction sets while maintaining the compelling accuracy guarantees. 
In this work, we follow the expanded admission CP extension of \citet{fisch2021efficient} that leverages the existence of equally correct answers to improve the statistical efficiency of the calibration. We refer the reader to \citet{fisch2021efficient} for the description and theoretical analysis of the method, and detail the exact setting below.

We use the top 20 predictions of the Luke model on \squad as candidate answers per question and use the conformal-cascades repository for running the calibration experiments.\footnote{\url{https://github.com/ajfisch/conformal-cascades}} We only utilize the expanded admission functionality of the code, and don't use a cascade here as we directly use the negative span score from Luke as the nonconformity measure. We run 50 calibration trials and report the average results, as well as visualize the 16 and 84th percentiles in Figure~\ref{fig:conformal}. In each trial, we randomly partition the data into 80\% calibration and 20\% test examples. Reference results for different target accuries are provided in Table~\ref{tab:conformal_results}.

CP prediction sets are computed as a function of the calibration examples and a user defined target accuracy. Following ~\citet{fisch2021efficient} we empirically verify the validity of the sets (i.e., marginally meeting the target accuracy), and measure the size of the sets. The goal is to minimize the size of the prediction sets while satisfying validity. Our experiments evaluate different equivalence terms for the admission expansion function.

For exact expanded admission, we experiment with either using the original \squad labels, or including our AE annotations. For approximate expanded admission, we try both \fone and our \ourscore equivalence metrics. We follow the method in Appendix B of ~\citet{fisch2021efficient} to statistically correct for the approximation errors of the metrics. We use only 10\% of the calibration data to compute the empirical FPR of the metric (i.e., the ratio that the top answer that was accepted by the metric was incorrect). We assume that the rest of the calibration data is lacking AE labels, hence the need for the approximation. We use binomial confidence intervals to get an upper bound of the true error, resulting in a two-level probabilistic guarantee.\footnote{We use scipy.special.betaincinv to compute the bound with $\gamma=0.01$.} Then, we apply the correction by dividing the p-value of each candidate by the lower bound of the TPR.

%% file: ae.bbl
\begin{thebibliography}{47}
\expandafter\ifx\csname natexlab\endcsname\relax\def\natexlab#1{#1}\fi

\bibitem[{Angelopoulos and Bates(2021)}]{angelopoulos2021gentle}
Anastasios~N. Angelopoulos and Stephen Bates. 2021.
\newblock \href {http://arxiv.org/abs/2107.07511} {A gentle introduction to
  conformal prediction and distribution-free uncertainty quantification}.

\bibitem[{Banerjee and Lavie(2005)}]{banerjee-lavie-2005-meteor}
Satanjeev Banerjee and Alon Lavie. 2005.
\newblock \href {https://aclanthology.org/W05-0909} {{METEOR}: An automatic
  metric for {MT} evaluation with improved correlation with human judgments}.
\newblock In \emph{Proceedings of the {ACL} Workshop on Intrinsic and Extrinsic
  Evaluation Measures for Machine Translation and/or Summarization}, pages
  65--72, Ann Arbor, Michigan. Association for Computational Linguistics.

\bibitem[{Bates et~al.(2021)Bates, Angelopoulos, Lei, Malik, and
  Jordan}]{bates2021distribution}
Stephen Bates, Anastasios Angelopoulos, Lihua Lei, Jitendra Malik, and Michael
  Jordan. 2021.
\newblock Distribution-free, risk-controlling prediction sets.
\newblock \emph{Journal of the ACM (JACM)}, 68(6):1--34.

\bibitem[{Bollacker et~al.(2008)Bollacker, Evans, Paritosh, Sturge, and
  Taylor}]{freebase08}
Kurt Bollacker, Colin Evans, Praveen Paritosh, Tim Sturge, and Jamie Taylor.
  2008.
\newblock \href {https://doi.org/10.1145/1376616.1376746} {Freebase: A
  collaboratively created graph database for structuring human knowledge}.
\newblock In \emph{Proceedings of the 2008 ACM SIGMOD International Conference
  on Management of Data}, SIGMOD '08, page 1247–1250, New York, NY, USA.
  Association for Computing Machinery.

\bibitem[{Breck et~al.(1999)Breck, Burger, Ferro, House, Light, and
  Mani}]{breck-1999-aw}
Eric Breck, John Burger, Lisa Ferro, David House, Marc Light, and Inderjeet
  Mani. 1999.
\newblock A sys called qanda.
\newblock In \emph{Proceedings of The Eighth Text {REtrieval} Conference}.

\bibitem[{Chen et~al.(2019)Chen, Stanovsky, Singh, and
  Gardner}]{chen-etal-2019-evaluating}
Anthony Chen, Gabriel Stanovsky, Sameer Singh, and Matt Gardner. 2019.
\newblock \href {https://doi.org/10.18653/v1/D19-5817} {Evaluating question
  answering evaluation}.
\newblock In \emph{Proceedings of the 2nd Workshop on Machine Reading for
  Question Answering}, pages 119--124, Hong Kong, China. Association for
  Computational Linguistics.

\bibitem[{Chen et~al.(2020)Chen, Stanovsky, Singh, and
  Gardner}]{chen-et-al-2020}
Anthony Chen, Gabriel Stanovsky, Sameer Singh, and Matt Gardner. 2020.
\newblock \href {https://doi.org/10.18653/v1/2020.emnlp-main.528} {{MOCHA:} {A}
  dataset for training and evaluating generative reading comprehension
  metrics}.
\newblock In \emph{Proceedings of the 2020 Conference on Empirical Methods in
  Natural Language Processing, {EMNLP} 2020, Online, November 16-20, 2020},
  pages 6521--6532. Association for Computational Linguistics.

\bibitem[{Chen et~al.(2021)Chen, Choi, and Durrett}]{chen-etal-2021-pw}
Jifan Chen, Eunsol Choi, and Greg Durrett. 2021.
\newblock \href {https://aclanthology.org/2021.findings-emnlp.324} {Can {NLI}
  models verify {QA} systems{'} predictions?}
\newblock In \emph{Findings of the Association for Computational Linguistics:
  EMNLP 2021}, pages 3841--3854, Punta Cana, Dominican Republic. Association
  for Computational Linguistics.

\bibitem[{Clark et~al.(2019)Clark, Celikyilmaz, and Smith}]{clark-etal-2019}
Elizabeth Clark, Asli Celikyilmaz, and Noah~A. Smith. 2019.
\newblock \href {https://doi.org/10.18653/v1/P19-1264} {Sentence mover{'}s
  similarity: Automatic evaluation for multi-sentence texts}.
\newblock In \emph{Proceedings of the 57th Annual Meeting of the Association
  for Computational Linguistics}, pages 2748--2760, Florence, Italy.
  Association for Computational Linguistics.

\bibitem[{Cohen et~al.(2003)Cohen, Ravikumar, and Fienberg}]{Cohen2003ACO}
William~W. Cohen, Pradeep Ravikumar, and Stephen~E. Fienberg. 2003.
\newblock A comparison of string distance metrics for name-matching tasks.
\newblock In \emph{IIWeb}.

\bibitem[{Devlin et~al.(2019)Devlin, Chang, Lee, and Toutanova}]{devlin-19}
Jacob Devlin, Ming-Wei Chang, Kenton Lee, and Kristina Toutanova. 2019.
\newblock \href {https://doi.org/10.18653/v1/N19-1423} {{BERT}: Pre-training of
  deep bidirectional transformers for language understanding}.
\newblock In \emph{Proceedings of the 2019 Conference of the North {A}merican
  Chapter of the Association for Computational Linguistics: Human Language
  Technologies, Volume 1 (Long and Short Papers)}, pages 4171--4186,
  Minneapolis, Minnesota. Association for Computational Linguistics.

\bibitem[{Elazar and Goldberg(2019)}]{yanai-2019-fused-head}
Yanai Elazar and Yoav Goldberg. 2019.
\newblock \href {https://doi.org/10.1162/tacl_a_00280} {{Where’s My Head?
  Definition, Data Set, and Models for Numeric Fused-Head Identification and
  Resolution}}.
\newblock \emph{Transactions of the Association for Computational Linguistics},
  7:519--535.

\bibitem[{Eyal et~al.(2019)Eyal, Baumel, and Elhadad}]{eyal-etal-2019-question}
Matan Eyal, Tal Baumel, and Michael Elhadad. 2019.
\newblock \href {https://doi.org/10.18653/v1/N19-1395} {Question answering as
  an automatic evaluation metric for news article summarization}.
\newblock In \emph{Proceedings of the 2019 Conference of the North {A}merican
  Chapter of the Association for Computational Linguistics: Human Language
  Technologies, Volume 1 (Long and Short Papers)}, pages 3938--3948,
  Minneapolis, Minnesota. Association for Computational Linguistics.

\bibitem[{Fabbri et~al.(2021)Fabbri, Wu, Liu, and Xiong}]{alex2021qafacteval}
Alexander~R. Fabbri, Chien-Sheng Wu, Wenhao Liu, and Caiming Xiong. 2021.
\newblock \href {http://arxiv.org/abs/2112.08542} {Qafacteval: Improved
  qa-based factual consistency evaluation for summarization}.
\newblock \emph{CoRR}, abs/2112.08542.

\bibitem[{Fisch et~al.(2022)Fisch, Jia, and Schuster}]{cp_tutorial2}
Adam Fisch, Robin Jia, and Tal Schuster. 2022.
\newblock Uncertainty estimation for natural language processing.
\newblock In \emph{COLING}.

\bibitem[{Fisch et~al.(2021)Fisch, Schuster, Jaakkola, and
  Barzilay}]{fisch2021efficient}
Adam Fisch, Tal Schuster, Tommi~S. Jaakkola, and Regina Barzilay. 2021.
\newblock \href {https://openreview.net/forum?id=tnSo6VRLmT} {Efficient
  conformal prediction via cascaded inference with expanded admission}.
\newblock In \emph{International Conference on Learning Representations}.

\bibitem[{Gehrmann et~al.(2021)Gehrmann, Adewumi, Aggarwal, Ammanamanchi,
  Anuoluwapo, Bosselut, Chandu, Clinciu, Das, Dhole, Du, Durmus, Dušek,
  Emezue, Gangal, Garbacea, Hashimoto, Hou, Jernite, Jhamtani, Ji, Jolly, Kale,
  Kumar, Ladhak, Madaan, Maddela, Mahajan, Mahamood, Majumder, Martins,
  McMillan-Major, Mille, van Miltenburg, Nadeem, Narayan, Nikolaev, Niyongabo,
  Osei, Parikh, Perez-Beltrachini, Rao, Raunak, Rodriguez, Santhanam, Sedoc,
  Sellam, Shaikh, Shimorina, Cabezudo, Strobelt, Subramani, Xu, Yang, Yerukola,
  and Zhou}]{gembenchmark2021}
Sebastian Gehrmann, Tosin Adewumi, Karmanya Aggarwal, Pawan~Sasanka
  Ammanamanchi, Aremu Anuoluwapo, Antoine Bosselut, Khyathi~Raghavi Chandu,
  Miruna Clinciu, Dipanjan Das, Kaustubh~D. Dhole, Wanyu Du, Esin Durmus,
  Ondřej Dušek, Chris Emezue, Varun Gangal, Cristina Garbacea, Tatsunori
  Hashimoto, Yufang Hou, Yacine Jernite, Harsh Jhamtani, Yangfeng Ji, Shailza
  Jolly, Mihir Kale, Dhruv Kumar, Faisal Ladhak, Aman Madaan, Mounica Maddela,
  Khyati Mahajan, Saad Mahamood, Bodhisattwa~Prasad Majumder, Pedro~Henrique
  Martins, Angelina McMillan-Major, Simon Mille, Emiel van Miltenburg, Moin
  Nadeem, Shashi Narayan, Vitaly Nikolaev, Rubungo~Andre Niyongabo, Salomey
  Osei, Ankur Parikh, Laura Perez-Beltrachini, Niranjan~Ramesh Rao, Vikas
  Raunak, Juan~Diego Rodriguez, Sashank Santhanam, João Sedoc, Thibault
  Sellam, Samira Shaikh, Anastasia Shimorina, Marco Antonio~Sobrevilla
  Cabezudo, Hendrik Strobelt, Nishant Subramani, Wei Xu, Diyi Yang, Akhila
  Yerukola, and Jiawei Zhou. 2021.
\newblock \href {http://arxiv.org/abs/2102.01672} {The gem benchmark: Natural
  language generation, its evaluation and metrics}.

\bibitem[{Gneezy and Rustichini(2000)}]{payenough}
Uri Gneezy and Aldo Rustichini. 2000.
\newblock \href {http://www.jstor.org/stable/2586896} {Pay enough or don't pay
  at all}.
\newblock \emph{The Quarterly Journal of Economics}, 115(3):791--810.

\bibitem[{Honovich et~al.(2022)Honovich, Aharoni, Herzig, Taitelbaum,
  Kukliansy, Cohen, Scialom, Szpektor, Hassidim, and
  Matias}]{honovich-etal-2022-true-evaluating}
Or~Honovich, Roee Aharoni, Jonathan Herzig, Hagai Taitelbaum, Doron Kukliansy,
  Vered Cohen, Thomas Scialom, Idan Szpektor, Avinatan Hassidim, and Yossi
  Matias. 2022.
\newblock \href {https://doi.org/10.18653/v1/2022.naacl-main.287} {{TRUE}:
  Re-evaluating factual consistency evaluation}.
\newblock In \emph{Proceedings of the 2022 Conference of the North American
  Chapter of the Association for Computational Linguistics: Human Language
  Technologies}, pages 3905--3920, Seattle, United States. Association for
  Computational Linguistics.

\bibitem[{Honovich et~al.(2021)Honovich, Choshen, Aharoni, Neeman, Szpektor,
  and Abend}]{honovich-etal-2021-q2}
Or~Honovich, Leshem Choshen, Roee Aharoni, Ella Neeman, Idan Szpektor, and Omri
  Abend. 2021.
\newblock \href {https://aclanthology.org/2021.emnlp-main.619} {$q^{2}$:
  {E}valuating factual consistency in knowledge-grounded dialogues via question
  generation and question answering}.
\newblock In \emph{Proceedings of the 2021 Conference on Empirical Methods in
  Natural Language Processing}, pages 7856--7870, Online and Punta Cana,
  Dominican Republic. Association for Computational Linguistics.

\bibitem[{Kocmi et~al.(2021)Kocmi, Federmann, Grundkiewicz, Junczys-Dowmunt,
  Matsushita, and Menezes}]{kocmi2021ship}
Tom Kocmi, Christian Federmann, Roman Grundkiewicz, Marcin Junczys-Dowmunt,
  Hitokazu Matsushita, and Arul Menezes. 2021.
\newblock \href {http://arxiv.org/abs/2107.10821} {To ship or not to ship: An
  extensive evaluation of automatic metrics for machine translation}.

\bibitem[{Krippendorff(2004)}]{krippendorff04}
Klaus Krippendorff. 2004.
\newblock \emph{Content Analysis: An Introduction to Its Methodology (second
  edition)}.
\newblock Sage Publications.

\bibitem[{Kusner et~al.(2015)Kusner, Sun, Kolkin, and
  Weinberger}]{kusner-etal-2015-ls}
Matt~J Kusner, Yu~Sun, Nicholas~I Kolkin, and Kilian~Q Weinberger. 2015.
\newblock From word embeddings to document distances.
\newblock In \emph{Proceedings of the 32nd International Conference on
  International Conference on Machine Learning - Volume 37}, ICML'15, pages
  957--966. JMLR.org.

\bibitem[{Lan et~al.(2019)Lan, Chen, Goodman, Gimpel, Sharma, and
  Soricut}]{albert-paper-19}
Zhenzhong Lan, Mingda Chen, Sebastian Goodman, Kevin Gimpel, Piyush Sharma, and
  Radu Soricut. 2019.
\newblock \href {http://arxiv.org/abs/1909.11942} {{ALBERT:} {A} lite {BERT}
  for self-supervised learning of language representations}.
\newblock \emph{CoRR}, abs/1909.11942.

\bibitem[{Lee et~al.(2019)Lee, Chang, and Toutanova}]{lee-etal-2019-latent}
Kenton Lee, Ming-Wei Chang, and Kristina Toutanova. 2019.
\newblock \href {https://doi.org/10.18653/v1/P19-1612} {Latent retrieval for
  weakly supervised open domain question answering}.
\newblock In \emph{Proceedings of the 57th Annual Meeting of the Association
  for Computational Linguistics}, pages 6086--6096, Florence, Italy.
  Association for Computational Linguistics.

\bibitem[{Lin(2004)}]{lin-2004-rouge}
Chin-Yew Lin. 2004.
\newblock \href {https://aclanthology.org/W04-1013} {{ROUGE}: A package for
  automatic evaluation of summaries}.
\newblock In \emph{Text Summarization Branches Out}, pages 74--81, Barcelona,
  Spain. Association for Computational Linguistics.

\bibitem[{Min et~al.(2020{\natexlab{a}})Min, Boyd{-}Graber, Alberti, Chen,
  Choi, Collins, Guu, Hajishirzi, Lee, Palomaki, Raffel, Roberts, Kwiatkowski,
  Lewis, Wu, K{\"{u}}ttler, Liu, Minervini, Stenetorp, Riedel, Yang, Seo,
  Izacard, Petroni, Hosseini, Cao, Grave, Yamada, Shimaoka, Suzuki, Miyawaki,
  Sato, Takahashi, Suzuki, Fajcik, Docekal, Ondrej, Smrz, Cheng, Shen, Liu, He,
  Chen, Gao, Oguz, Chen, Karpukhin, Peshterliev, Okhonko, Schlichtkrull, Gupta,
  Mehdad, and Yih}]{min-et-al-2021}
Sewon Min, Jordan~L. Boyd{-}Graber, Chris Alberti, Danqi Chen, Eunsol Choi,
  Michael Collins, Kelvin Guu, Hannaneh Hajishirzi, Kenton Lee, Jennimaria
  Palomaki, Colin Raffel, Adam Roberts, Tom Kwiatkowski, Patrick S.~H. Lewis,
  Yuxiang Wu, Heinrich K{\"{u}}ttler, Linqing Liu, Pasquale Minervini, Pontus
  Stenetorp, Sebastian Riedel, Sohee Yang, Minjoon Seo, Gautier Izacard, Fabio
  Petroni, Lucas Hosseini, Nicola~De Cao, Edouard Grave, Ikuya Yamada, Sonse
  Shimaoka, Masatoshi Suzuki, Shumpei Miyawaki, Shun Sato, Ryo Takahashi, Jun
  Suzuki, Martin Fajcik, Martin Docekal, Karel Ondrej, Pavel Smrz, Hao Cheng,
  Yelong Shen, Xiaodong Liu, Pengcheng He, Weizhu Chen, Jianfeng Gao, Barlas
  Oguz, Xilun Chen, Vladimir Karpukhin, Stan Peshterliev, Dmytro Okhonko,
  Michael~Sejr Schlichtkrull, Sonal Gupta, Yashar Mehdad, and Wen{-}tau Yih.
  2020{\natexlab{a}}.
\newblock \href {http://proceedings.mlr.press/v133/min21a.html} {Neurips 2020
  efficientqa competition: Systems, analyses and lessons learned}.
\newblock In \emph{NeurIPS 2020 Competition and Demonstration Track, 6-12
  December 2020, Virtual Event / Vancouver, BC, Canada}, volume 133 of
  \emph{Proceedings of Machine Learning Research}, pages 86--111. {PMLR}.

\bibitem[{Min et~al.(2020{\natexlab{b}})Min, Michael, Hajishirzi, and
  Zettlemoyer}]{min-et-al-2020}
Sewon Min, Julian Michael, Hannaneh Hajishirzi, and Luke Zettlemoyer.
  2020{\natexlab{b}}.
\newblock \href {http://arxiv.org/abs/2004.10645} {Ambigqa: Answering ambiguous
  open-domain questions}.
\newblock \emph{CoRR}, abs/2004.10645.

\bibitem[{Papineni et~al.(2002)Papineni, Roukos, Ward, and Zhu}]{bleu}
Kishore Papineni, Salim Roukos, Todd Ward, and Wei-Jing Zhu. 2002.
\newblock Bleu: A method for automatic evaluation of machine translation.
\newblock In \emph{Proceedings of the 40th Annual Meeting on Association for
  Computational Linguistics}, ACL '02, page 311–318, USA. Association for
  Computational Linguistics.

\bibitem[{Rajpurkar et~al.(2016)Rajpurkar, Zhang, Lopyrev, and
  Liang}]{rajpurkar-16}
Pranav Rajpurkar, Jian Zhang, Konstantin Lopyrev, and Percy Liang. 2016.
\newblock \href {https://doi.org/10.18653/v1/D16-1264} {{SQ}u{AD}: 100,000+
  questions for machine comprehension of text}.
\newblock In \emph{Proceedings of the 2016 Conference on Empirical Methods in
  Natural Language Processing}, pages 2383--2392, Austin, Texas. Association
  for Computational Linguistics.

\bibitem[{Rei et~al.(2020)Rei, Stewart, Farinha, and
  Lavie}]{rei-etal-2020-comet}
Ricardo Rei, Craig Stewart, Ana~C Farinha, and Alon Lavie. 2020.
\newblock \href {https://doi.org/10.18653/v1/2020.emnlp-main.213} {{COMET}: A
  neural framework for {MT} evaluation}.
\newblock In \emph{Proceedings of the 2020 Conference on Empirical Methods in
  Natural Language Processing (EMNLP)}, pages 2685--2702, Online. Association
  for Computational Linguistics.

\bibitem[{Risch et~al.(2021)Risch, Möller, Gutsch, and
  Pietsch}]{risch2021semantic}
Julian Risch, Timo Möller, Julian Gutsch, and Malte Pietsch. 2021.
\newblock \href {http://arxiv.org/abs/2108.06130} {Semantic answer similarity
  for evaluating question answering models}.

\bibitem[{Schuster et~al.(2021{\natexlab{a}})Schuster, Fisch, and
  Barzilay}]{schuster-etal-2021-get}
Tal Schuster, Adam Fisch, and Regina Barzilay. 2021{\natexlab{a}}.
\newblock \href {https://doi.org/10.18653/v1/2021.naacl-main.52} {Get your
  vitamin {C}! robust fact verification with contrastive evidence}.
\newblock In \emph{Proceedings of the 2021 Conference of the North American
  Chapter of the Association for Computational Linguistics: Human Language
  Technologies}, pages 624--643, Online. Association for Computational
  Linguistics.

\bibitem[{Schuster et~al.(2022)Schuster, Fisch, Gupta, Dehghani, Bahri, Tran,
  Tay, and Metzler}]{Schuster2022CALM}
Tal Schuster, Adam Fisch, Jai Gupta, Mostafa Dehghani, Dara Bahri, Vinh~Quang
  Tran, Yi~Tay, and Donald Metzler. 2022.
\newblock \href {https://arxiv.org/abs/2207.07061} {Confident adaptive language
  modeling}.
\newblock In \emph{Advances in Neural Information Processing Systems
  (NeurIPS)}.

\bibitem[{Schuster et~al.(2021{\natexlab{b}})Schuster, Fisch, Jaakkola, and
  Barzilay}]{schuster-etal-2021-consistent}
Tal Schuster, Adam Fisch, Tommi Jaakkola, and Regina Barzilay.
  2021{\natexlab{b}}.
\newblock \href {https://aclanthology.org/2021.emnlp-main.406} {Consistent
  accelerated inference via confident adaptive transformers}.
\newblock In \emph{Proceedings of the 2021 Conference on Empirical Methods in
  Natural Language Processing}, pages 4962--4979, Online and Punta Cana,
  Dominican Republic. Association for Computational Linguistics.

\bibitem[{Sellam et~al.(2020{\natexlab{a}})Sellam, Das, and
  Parikh}]{sellam-etal-2020-bleurt}
Thibault Sellam, Dipanjan Das, and Ankur Parikh. 2020{\natexlab{a}}.
\newblock \href {https://doi.org/10.18653/v1/2020.acl-main.704} {{BLEURT}:
  Learning robust metrics for text generation}.
\newblock In \emph{Proceedings of the 58th Annual Meeting of the Association
  for Computational Linguistics}, pages 7881--7892, Online. Association for
  Computational Linguistics.

\bibitem[{Sellam et~al.(2020{\natexlab{b}})Sellam, Pu, Chung, Gehrmann, Tan,
  Freitag, Das, and Parikh}]{sellam-etal-2020-learning}
Thibault Sellam, Amy Pu, Hyung~Won Chung, Sebastian Gehrmann, Qijun Tan, Markus
  Freitag, Dipanjan Das, and Ankur Parikh. 2020{\natexlab{b}}.
\newblock \href {https://aclanthology.org/2020.wmt-1.102} {Learning to evaluate
  translation beyond {E}nglish: {BLEURT} submissions to the {WMT} metrics 2020
  shared task}.
\newblock In \emph{Proceedings of the Fifth Conference on Machine Translation},
  pages 921--927, Online. Association for Computational Linguistics.

\bibitem[{Seo et~al.(2016)Seo, Kembhavi, Farhadi, and
  Hajishirzi}]{bidaf-paper-16}
Min~Joon Seo, Aniruddha Kembhavi, Ali Farhadi, and Hannaneh Hajishirzi. 2016.
\newblock \href {http://arxiv.org/abs/1611.01603} {Bidirectional attention flow
  for machine comprehension}.
\newblock \emph{CoRR}, abs/1611.01603.

\bibitem[{Shafer and Vovk(2007)}]{cp_tutorial}
Glenn Shafer and Vladimir Vovk. 2007.
\newblock \href {http://arxiv.org/abs/0706.3188} {A tutorial on conformal
  prediction}.
\newblock \emph{CoRR}, abs/0706.3188.

\bibitem[{Si et~al.(2021)Si, Zhao, and Boyd-Graber}]{Si:Zhao:Boyd-Graber-2021}
Chenglei Si, Chen Zhao, and Jordan Boyd-Graber. 2021.
\newblock \href {http://umiacs.umd.edu/~jbg//docs/2021_emnlp_answer_equiv.pdf}
  {What’s in a name? answer equivalence for open-domain question answering}.
\newblock In \emph{Empirical Methods in Natural Language Processing}.

\bibitem[{Voorhees and Tice(2000)}]{vorhees2000}
Ellen~M. Voorhees and Dawn~M. Tice. 2000.
\newblock \href {https://doi.org/10.1145/345508.345577} {Building a question
  answering test collection}.
\newblock In \emph{{SIGIR} 2000: Proceedings of the 23rd Annual International
  {ACM} {SIGIR} Conference on Research and Development in Information
  Retrieval, July 24-28, 2000, Athens, Greece}, pages 200--207. {ACM}.

\bibitem[{Vovk et~al.(2005)Vovk, Gammerman, and Shafer}]{vovk_random}
Vladimir Vovk, Alex Gammerman, and Glenn Shafer. 2005.
\newblock \emph{Algorithmic Learning in a Random World}.
\newblock Springer-Verlag, Berlin, Heidelberg.

\bibitem[{Weston et~al.(2015)Weston, Bordes, Chopra, Rush, van Merriënboer,
  Joulin, and Mikolov}]{weston2015aicomplete}
Jason Weston, Antoine Bordes, Sumit Chopra, Alexander~M. Rush, Bart van
  Merriënboer, Armand Joulin, and Tomas Mikolov. 2015.
\newblock \href {http://arxiv.org/abs/1502.05698} {Towards ai-complete question
  answering: A set of prerequisite toy tasks}.
\newblock \emph{CoRR}.

\bibitem[{Yamada et~al.(2020)Yamada, Asai, Shindo, Takeda, and
  Matsumoto}]{luke-paper-16}
Ikuya Yamada, Akari Asai, Hiroyuki Shindo, Hideaki Takeda, and Yuji Matsumoto.
  2020.
\newblock \href {https://doi.org/10.18653/v1/2020.emnlp-main.523} {{LUKE:} deep
  contextualized entity representations with entity-aware self-attention}.
\newblock In \emph{Proceedings of the 2020 Conference on Empirical Methods in
  Natural Language Processing, {EMNLP} 2020, Online, November 16-20, 2020},
  pages 6442--6454. Association for Computational Linguistics.

\bibitem[{Yang et~al.(2018)Yang, Liu, Liu, Lyu, and
  Li}]{yang-etal-2018-adaptations}
An~Yang, Kai Liu, Jing Liu, Yajuan Lyu, and Sujian Li. 2018.
\newblock \href {https://doi.org/10.18653/v1/W18-2611} {Adaptations of {ROUGE}
  and {BLEU} to better evaluate machine reading comprehension task}.
\newblock In \emph{Proceedings of the Workshop on Machine Reading for Question
  Answering}, pages 98--104, Melbourne, Australia. Association for
  Computational Linguistics.

\bibitem[{Yang et~al.(2019)Yang, Dai, Yang, Carbonell, Salakhutdinov, and
  Le}]{yang-xlnet}
Zhilin Yang, Zihang Dai, Yiming Yang, Jaime Carbonell, Russ~R Salakhutdinov,
  and Quoc~V Le. 2019.
\newblock \href
  {https://proceedings.neurips.cc/paper/2019/file/dc6a7e655d7e5840e66733e9ee67cc69-Paper.pdf}
  {Xlnet: Generalized autoregressive pretraining for language understanding}.
\newblock In \emph{Advances in Neural Information Processing Systems},
  volume~32. Curran Associates, Inc.

\bibitem[{Zhang et~al.(2020)Zhang, Kishore, Wu, Weinberger, and
  Artzi}]{bertscore}
Tianyi Zhang, Varsha Kishore, Felix Wu, Kilian~Q. Weinberger, and Yoav Artzi.
  2020.
\newblock Bertscore: Evaluating text generation with {BERT}.
\newblock In \emph{8th International Conference on Learning Representations,
  {ICLR} 2020, Addis Ababa, Ethiopia, April 26-30, 2020}.

\end{thebibliography}
